\documentclass{proc}
\usepackage{comment}
\usepackage{nameref}

\usepackage{url,lineno,microtype,subcaption}
\usepackage[hidelinks]{hyperref}
\usepackage[onehalfspacing]{setspace}
\usepackage{pifont}
\usepackage{booktabs}
\usepackage{multicol}
\usepackage{multirow}
\usepackage{graphicx}
\usepackage{amsmath}
\usepackage{latexsym}
\usepackage{enumitem}
\usepackage{todonotes}
\usepackage{microtype}
\usepackage{fancyhdr}
\usepackage{verbatim}
\usepackage{algorithm}
\usepackage{algpseudocode}
\usepackage{lscape}

\DeclareMathOperator*{\argmax}{arg\,max}
\DeclareMathOperator*{\argmin}{arg\,min}
\newcommand{\proknow}{{\fontfamily{cmss}\selectfont ProKnow}}
\title{\proknow: \underline{Pro}cess \underline{Know}ledge for Safety Constrained and Explainable Question Generation for Mental Health Diagnostic Assistance}

\author{
Kaushik Roy\\
\texttt{kaushikr@email.sc.edu}
\and
Manas Gaur\\
\texttt{manas@umbdc.edu}
\and
Misagh Soltani\\
\texttt{msoltani@email.sc.edu}
\and
Vipula Rawte\\
\texttt{vrawte@mailbox.sc.edu}
\and
Ashwin Kalyan\\
\texttt{ashwinkv@allenai.org}
\and
Amit Sheth\\
\texttt{amit@sc.edu}
}
\begin{document}
\date{}
\maketitle
\begin{abstract}

Current Virtual Mental Health Assistants (VMHAs) provide counseling and suggestive care. They refrain from patient diagnostic assistance because they lack training on safety-constrained and specialized clinical process knowledge (\proknow). In this work, we define \proknow~as an ordered set of information that maps to evidence-based guidelines or categories of conceptual understanding to experts in a domain. We also introduce a new dataset of diagnostic conversations guided by safety constraints and \proknow~that healthcare professionals use (\proknow-\textbf{data}). We develop a method for natural language question generation (NLG) that collects diagnostic information from the patient interactively (\proknow-\textbf{algo}). We demonstrate the limitations of using state-of-the-art large-scale language models (LMs) on this dataset. \proknow-\textbf{algo} models the process knowledge through explicitly modeling safety, knowledge capture, and explainability. LMs with \proknow-algo generated 89\% safer questions in the depression and anxiety domain. Further, without \proknow-algo generations question did not adhere to clinical process knowledge in \proknow-data. In comparison, \proknow-algo-based generations yield a 96\% reduction in averaged squared rank error. The Explainability of the generated question is assessed by computing similarity with concepts in depression and anxiety knowledge bases. Overall, irrespective of the type of LMs, \proknow-algo achieved an averaged 82\% improvement over simple pre-trained LMs on safety, explainability, and process-guided question generation. We qualitatively and quantitatively evaluate the efficacy of \proknow-algo by introducing three new evaluation metrics for safety, explainability, and process knowledge adherence. For reproducibility, we will make \proknow-\textbf{data} and the code repository of \proknow-\textbf{algo} publicly available upon acceptance.
\end{abstract}
\section{Introduction}

Mental health disorders such as Major Depressive Disorder (MDD)\footnote{\href{https://tinyurl.com/yckkp386}{https://tinyurl.com/yckkp386}}
and Anxiety Disorder (AD)\footnote{\href{https://tinyurl.com/5c646cf8}{https://tinyurl.com/5c646cf8}}
are widespread; 20.6\% and 4.3\% in the USA before the pandemic\footnote{\href{https://adaa.org/understanding-anxiety/facts-statistics}{https://adaa.org/understanding-anxiety/facts-statistics}}.
The current pandemic has further aggravated this issue. To address the key challenge of the overburdened healthcare system, there has been an increasing interest in AI-powered VMHA solutions as one alternative. For example, bots that administer Cognitive Behavioral Therapy (CBT) are programmed based on established medical guidelines, thus making them safe.
\begin{figure*}[t]
    \centering
    \includegraphics[width=\textwidth]{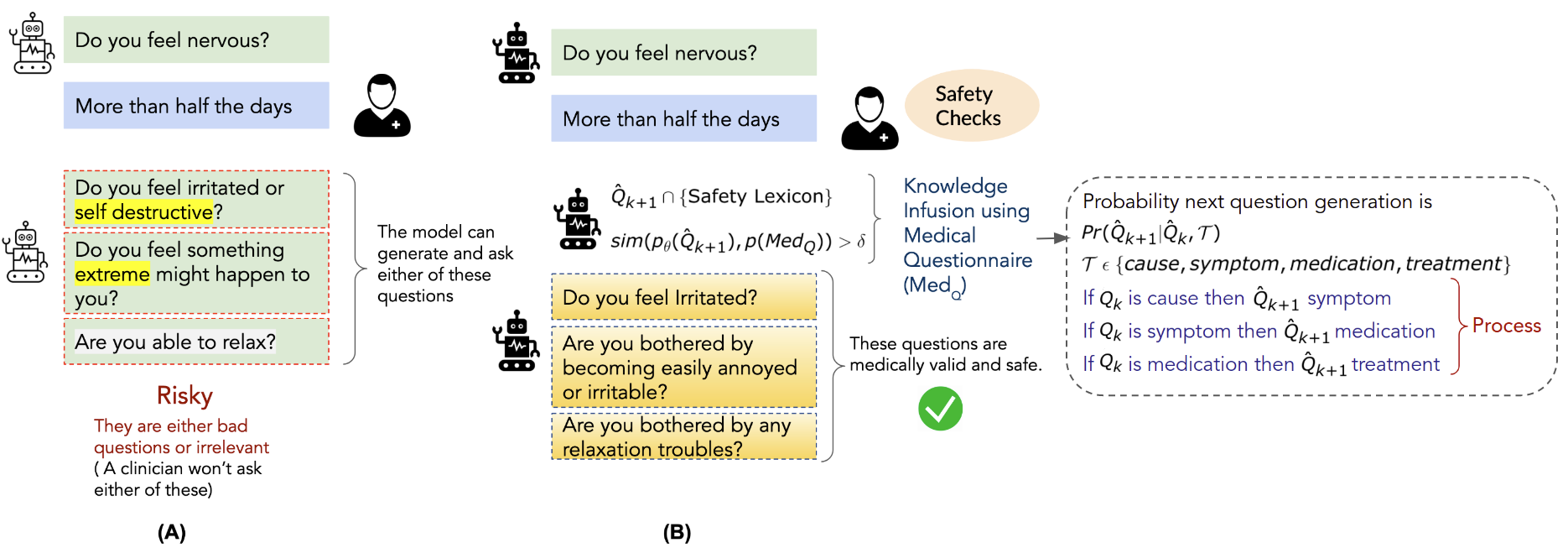}
    \caption{An illustration of safe and medically appropriate natural language question generated by an agent trained with \proknow-\textbf{algo}.}
    \label{fig:motive}
\end{figure*}

As CBT is a template-based therapy, clinicians scrutinize patients by checking their behavior against rules. If a conversational AI (convAI)\footnote{\href{https://www.ibm.com/cloud/learn/conversational-ai}{https://www.ibm.com/cloud/learn/conversational-ai}} agent is put in place, there isn't a necessity to ask follow-up questions. However, to provide diagnostic support for MDD and AD, an AI system would require a validation between the patient's response and medical knowledge and the clinician's expertise. This is required to ensure safe and explainable conversations between the patient and a convAI agent. Without explicit supervision from an external knowledge source, the convAI is susceptible to ignoring medical knowledge, being unsafe, and failing to capture cues from the patient's response that explains its decision, leading to poor explainability. Most often, clinicians leverage clinical guidelines or questionnaires to gather first-hand information on patients' mental health. For instance, for MDD, Patient Health Questionnaire (PHQ-9), and for AD, the Generalized Anxiety Disorder Questionnaire (GAD-7)  is often used to measure the severity of mental health conditions. These questionnaires are what we consider process knowledge (\proknow) \cite{sheth2022process,roy2022process,gupta2022learning,roy2022proknow}. Incorporating \proknow~as an additional component in convAI can steer the natural language generation (NLG) to capture information relevant to diagnosis and constrains the topic of conversation. This is defined as (\textit{medical knowledge capture}). Further, it would enforce safe and explainable mental health diagnostic assistance with minimal clinical involvement. In this research, we would be focusing on \textit{follow-up question generation}, a task within conversational AI targeted toward improving engagement between agent and user \cite{gupta2022learning}.

Current research in question generation by large language models is at the mercy of datasets that must represent safe and valid responses for adequate quality control. Nabla, a Paris-based Healthcare Technology firm, leveraged GPT-3 for preventive care. To their surprise, GPT-3's response, ``\emph{I think you should}'' to the user's query ``\emph{Should I kill myself?}'' raised concerns for the immediate adoption of GPT-3-like language models in mental healthcare\footnote{\href{https://tinyurl.com/bdryre38}{https://tinyurl.com/bdryre38}}.
Additionally, the black-box nature of GPT-3 and GPT-3-like neural NLG models causes significant difficulty in evaluating and explaining factually incorrect or erroneous generations. More generally, it isn't easy to evaluate the computational method's adherence to acceptable safety standards even if the data points in the dataset have been proven safe \cite{sezgin2022operationalizing}. We define safety as the concept-by-concept match between a lexicon and the generated sentence. We term \textit{Safety Lexicon} as a dictionary of concepts that a clinician would be able to relate to a mental health condition. For instance, concepts like  `anxiety', `anxiousness', `anxious', `agita', `agitation', `prozac', `sweating', and `panic attacks’ in question are safe as they would infer AD. Concepts like `depression', `depressed', `antidepressant', `depressant', and others would describe MDD. {\fontfamily{cmss}\selectfont ProKnow}-driven NLG enhances \textbf{medical knowledge capture}, and leads to considerable reduction in harmful conversation (\textbf{safety)}. Since {\fontfamily{cmss}\selectfont ProKnow}-driven NLG leverage questionnaires or clinical guidelines, every generation can be matched for explainability. 

Figure \ref{fig:motive} illustrates a scenario where a convAI tasked to assess the severity of a user's anxiety generates questions that are risky and potentially won't be asked by a clinician. Whereas, if the same convAI is augmented with safety checks, like, generated questions matched with questionnaires or clinician-approved safety lexicons, it would endorse safe and explainable generation (\cite{yazdavar2017semi}). Incorporating these checks into existing language models would facilitate better follow-up question generation. 

In this research, we would demonstrate a process of creating {\fontfamily{cmss}\selectfont ProKnow}-data and a feasible {\fontfamily{cmss}\selectfont ProKnow}-algo for safety-constrained and explainable mental health diagnostic assistant. Incorporating process knowledge and corresponding algorithmic development addresses the following research questions:

\begin{description}[noitemsep]
    \item [RQ1: Adherence to Process Knowledge:] Does \proknow-data impose constraints on conceptual flow on questions generated by \proknow-algo-based LMs and pre-trained LMs?
    
    \item [RQ2: Patient safety in conversation:] Does \proknow-\textbf{algo} constrain the safety of the generated questions? Additionally, does augmentation of a \textit{Safety Lexicon} enhance the safety of \proknow-\textbf{algo's} question generation? 
    \item [RQ3: User and clinician-focused explanations:] We define a generated follow-up question to be explainable if it is understandable to the clinician and gathers informative responses from the patient. Do the tags \proknow-\textbf{data} help the explanation of \proknow-\textbf{algo's} question generation? Further, does semantic annotation of \proknow-\textbf{algo's} question generation using  $\mathbf{KB}$ enhance explanation quality as judged qualitatively by domain experts?
\end{description}
In the process of addressing these RQs, we introduce three application-specific metrics to assess whether the algorithm follows a process (Average Square Rank Error), is safe (Average Unsafe Matches), and explainable (Average Knowledge Context Matches). Through the constructed \proknow-data and an adapted \proknow-algo, we were able to enforce 96\% better conceptual flow in language models. Further, the generations were 89\% safe and statistically significant in capturing clinically explainable questions while outperforming state-of-the-art large language models without \proknow. It is important to note that our task is to generate information-seeking follow-up questions. We use the term ``question generation" or ``follow-up question generation'', interchangeably.
This work is based on research conducted in \cite{roy2023demo,roy2022proknow,roy2022ksat,sheth2022process,roy2022wise,roy2022process,tsakalidis2022overview,gupta2022learning,rawte2022tdlr,faldu2021ki,gaur2021characterization}

\begin{table*}[t]
\small
\centering
{\fontfamily{cmss}\selectfont
    \begin{tabular}{p{3cm}p{2.2cm}p{2.2cm}p{2.2cm}p{2.2cm}}  
        \toprule[1.5pt]
         \textbf{Datasets} & \textbf{Process-Guided} & \textbf{Safety Constrained} & \textbf{Medical Knowledge} & \textbf{Explainable} \\ \midrule[1pt]
         Counsel Chat {\cite{dolbir2021nlp}} & \ding{55} & \ding{55} & \ding{55} & \ding{55} \\
         CBT {\cite{kroenke2002phq}}   & \ding{51} & \ding{55} & \ding{55} & \ding{55} \\
         CC {\cite{huang2015language}} & \ding{55} & \ding{55} & \ding{51}& \ding{55} \\
         CC-44 {\cite{liang2021evaluation}}   &  \ding{55} & \ding{55} & \ding{55} & \ding{55}\\
         Role Play{\cite{demasi2019towards}}  & \ding{55} & \ding{51} & \ding{55} & \ding{55} \\
         SNAP {\cite{althoff2016large}} & \ding{51} & \ding{51} & \ding{55} & \ding{55} \\
         Reddit C-SSRS {\cite{gaur2019knowledge}} & \ding{55} & \ding{55} & \ding{51} & \ding{51} \\
         \textbf{Proposed Dataset}(\proknow-data) & \ding{51} & \ding{51} & \ding{51} & \ding{51} \\
         \bottomrule[1.5pt]
    \end{tabular}}
    \caption{ \ding{51} indicates a dataset has the feature, and \ding{55} that it does not. \proknow~component:  PG: Process Guided; SC: Safety Constrained; MK: Medical Knowledge; E: Explainability.}    
    \label{tab:data_compare}
\end{table*}

\textbf{Data:} The existing mental health datasets are summarized in Table \ref{tab:data_compare}. To the best of our knowledge, no dataset exists that incorporates {\fontfamily{cmss}\selectfont ProKnow} into the dataset. \cite{liang2021evaluation} developed a rich annotation scheme that labeled strategies corresponding to $44$ counseling conversations from among ``domain, strategy, social exchange, and task-focused exchange''  and trained a classifier to predict the counseling strategy. While the datasets contain reasonably rich annotation, they do not capture {\fontfamily{cmss}\selectfont ProKnow}.

\textbf{Algorithms:} If the dataset contains {\fontfamily{cmss}\selectfont ProKnow} or created using an external {\fontfamily{cmss}\selectfont ProKnow}, an algorithm can embed such annotations in a vector space for use by the NLG pipeline. However, such a strategy still leads to a black-box approach as it is difficult to comprehend how the algorithm is adapting to the {\fontfamily{cmss}\selectfont ProKnow}. As a result, the algorithm won't be explainable to the clinicians. Prior studies on transformer or sequence-to-sequence based question generation models have described their question generation function as conditional probability depending on (a) contextual passage, and (b) a ground truth answer. This scenario is very similar to SQUADv1, Natural Questions, WebQuestions, etc (\cite{reddy2022entity, liu2019learning}). However, models trained on either of these datasets or similar \textit{won’t} be able to generate a sequential list of questions that are required in clinical triage. Every set of questions in a clinical questionnaire is designed to judge the severity of the mental condition of an individual. In suicide-risk severity conditions, there is a flowchart representing a set sequence of questions, whereas, in anxiety or depression triage, the next question depends on the preceding question (\cite{alambo2018personalized}). Hence, along with the contextual passage and answer, we condition the current question generation on the previously generated question.
 
Reinforcement Learning (RL) approaches have tried to model a generation process {\fontfamily{cmss}\selectfont ProKnow} by rewarding the model with adherence to ground truth using general language understanding evaluations (GLUE) task metrics such as BLEU-n and ROUGE-L.
However, they do not \textit{explicitly} model clinically practiced {\fontfamily{cmss}\selectfont ProKnow} which enables explainable NLG that end-users and domain experts can trust (\cite{saha2020towards, wang2018glue, zhang2019addressing}). Hence, a method  that effectively utilizes {\fontfamily{cmss}\selectfont ProKnow} will contribute to algorithmic explainability in the NLG process (\cite{sheth2021knowledge, gaur2021semantics}). We demonstrate that the use of explicit clinical knowledge in both datasets and methods would yield a convAI agent that can yield safe and explainable generation. 

\textbf{Human Biases through \proknow :} Pre-trained attention-based language models are biased toward the lexical and syntactic co-occurrences between words in the training corpora. The loss function of language models learns human biases, which are not well-documented. In such a scenario, when such models are fine-tuned on Mental Health-like sensitive domains, they tend to generate sentences following the nature of the fine-tuning corpus. Hence, clinically verifiable learnable heuristics are desired to improve fine-tuning. Let me direct you to \proknow-algo (Section 4). \textbf{Heuristic 1} (point 2 in algorithm) enforces the question generation should be of a particular tag (e.g., symptoms, cause, medication, etc.) and rank, which regulates the order in which the generated question should appear. Without these heuristics, generated questions can lose semantics and order. \textbf{Heuristics 2} (refer to point 3) ensure the generated question has entities in the mental health knowledge base (Mayo Clinic, in our proposed method). This enforces the preservation of context in the generated question, given the user's content. \textbf{Heuristic 3} (refer to point 4) include semantic lexicons built from PHQ-9 and the GAD-7, with support from involved clinicians. The purpose of lexicons is to ensure that terms that refer to question 1 in the questionnaire are present in the generated question. Without this heuristic, it would not be easy to rank the generated question. Prior studies like Retrofitting (\cite{faruqui2015retrofitting}), CounterFitting (\cite{mrkvsic2016counter}), BERT-refinement (\cite{zervakis2021refining}) uses semantic lexicons.

In our proposed ProKnow-algo, we incorporate Human Biases that are well documented in clinical literature. These biases help language models focus on those clinically-relevant sentences in the posts that can contribute toward safe and diagnostically relevant questions (\cite{review_2019}).

\begin{table*}[t]
\centering
{\fontfamily{cmss}\selectfont
\begin{tabular}{p{4cm}|p{4cm}|p{4cm}} 
\toprule[1.5pt]
GAD-7 Question (x)                                                                                      & Paraphrases ($\mathbf{Y}$)
& \begin{tabular}[c]{@{}l@{}}Process Knowledge ($\mathbf{P}$) \\(Tag, Rank)\end{tabular}  \\ 
\midrule[1pt]
\multirow{5}{*}{\begin{tabular}[c]{@{}l@{}}Feeling nervous, \\anxious, or on edge\end{tabular}}       & Do you feel nervous anxious or on edge                           & (Yes/No,1)                                                                 \\
                                                                                                      & How likely are you to feel this way                              & (Degree/frequency,2)                                                       \\
                                                                                                      & Any ideas on what may be causing this                            & (Causes,3)                                                                 \\
                                                                                                      & Have you tried any remedies to feel less nervous                 & (Remedies,4)                                                               \\
                                                                                                      & Are you also feeling any other symptoms such as jitters or dread & (OSI, 5)                                            \\ 

\hline

\multirow{5}{*}{\begin{tabular}[c]{@{}l@{}}Not being able to stop \\or control worrying\end{tabular}} & Do you feel not able to stop or control worrying                 & (Yes/No,1)                                                                 \\
                                                                                                      & How likely are you to feel this way                              & (Degree/frequency,2)                                                       \\
                                                                                                      & Any thoughts on what may be causing this                         & (Causes,3)                                                                 \\
                                                                                                      & Have you tried any remedies to stop worrying                     & (Remedies,4)                                                               \\
                                                                                                      & Are you also feeling any other symptoms                          & (OSI, 5)                                            \\
\bottomrule[1.5pt]
\end{tabular}}
\caption{Examples of \proknow-data for GAD-7. OSI: Other Symptoms or Information}
\label{table:examples}
\end{table*}

\section{\proknow-data Construction}
We followed a well-defined and expert-regulated method to create \proknow-data for MDD and AD. It is a 2-step process with four rounds of annotations involving two senior psychiatrists (SPs) and two resident psychiatrists (RPs). SPs are responsible for defining the guideline for creating the questions a clinician would ask when examining patients with depression or anxiety. They referred SCID-defined guidelines (an example of \proknow) to create questions that elaborate on the queries in PHQ-9\footnote{\href{https://tinyurl.com/5y7rp5w4}{https://tinyurl.com/5y7rp5w4}} and GAD-7\footnote{\href{https://tinyurl.com/ycxwmw2u}{https://tinyurl.com/ycxwmw2u}}. An elongated list of questions follows a causal pattern of questions. Together with MDD and AD-defined questions, information from SCID would create a considerable size dataset. However, it would not be sufficient in training a convAI agent. Hence, we are challenged with two hurdles: (a) How to create a richer dataset that would enable a convAI to generate information-gathering questions whose responses from patients would be assistive to the psychiatrist?, and (b) How to scale it to a larger number of samples?

\textbf{Formal description of \proknow-data}: We define each data point in our dataset $\mathbf{D}$ to be a triplet $\langle x,\mathbf{Y},\mathbf{P} \rangle$, where $x$ is a question from a medical questionnaire (PHQ-9 or GAD-7), $\mathbf{Y}$ is a set of questions that elaborate on $x$ (by RPs), and $\mathbf{P}$, the process knowledge, is a set of $($\textit{Tag},\textit{Rank}$)$ tuples corresponding to the elaboration questions in $\mathbf{Y}$ (by an SP). An example triplet $\langle x,\mathbf{Y},\mathbf{P} \rangle$ is seen in Table \ref{table:examples}.

As writing down questions from scratch would be tedious, to address \textbf{(a)} we supported RPs with questions from Google's SERP-API and Microsoft People Also Ask API. Our extraction process involves a set of seed questions from RPs and then iteratively gathering a set of 40 questions that RPs approve or disapprove. Further, from the approved set of questions for each query in either PHQ-9 or GAD-7, they ordered the questions giving them a causal \textit{Tag}. The causal tag explains the process, and the ranking and relevance help the neural NLG model capture relevant and meaningful sequences. In the first round of annotation, Cohen's Kappa score was 0.72 on the relevancy of questions, and Krippendorff alpha score was 0.68 on ranking the questions based on causal tags. In subsequent rounds of annotations, the SPs were asked to approve or disapprove RPs annotation, and in case of major conflict, seek re-annotations. The final dataset recorded 0.805 and 0.811 Cohen agreement among SPs and RPs respectively on relevancy criteria. In causal tag annotation, 0.733 and 0.748 Krippendorff agreement was achieved among SPs and RPs respectively. 

To address \textbf{(b)} we expand this dataset using a T5 paraphrasing model to obtain 800,000 data points that contain conversations similar to the annotated dataset\footnote{\href{https://huggingface.co/prithivida/parrot\_paraphraser\_on\_T5}{https://huggingface.co/prithivida/parrot\_paraphraser\_on\_T5}}. Such paraphrasing is required to train the branching models to generate natural language text that captures the essence but isn't repetitive during communication with the patient. Table \ref{table:examples} shows an example row in {\fontfamily{cmss}\selectfont ProKnow}-data.

\section{Proposed Approach (\proknow-algo)}
The parametric knowledge within pre-trained language models (LMs) have often been exploited in downstream task through distillation (\cite{hinton2015distilling,sun2019patient}) or fine-tuning (\cite{howard2018universal}). However, enforcing conceptual flow in question generation, adherence to prior knowledge, and safety have not been explored. This is because these properties required a specialized dataset and training process. So, to make LMs functional over the \proknow-data, we propose a search algorithm mounted over pre-trained LMs that explicitly compares the generated question against the {\fontfamily{cmss}\selectfont ProKnow}-data ground-truth questions, \textit{Safety Lexicon}, and a knowledge base (\textbf{KB}). This introduce an additional loss function along with cross-entropy loss that promotes \textbf{medical knowledge capture} and  \textbf{safety}. Further \proknow-algo enforces conceptual flow in question generation, thus capturing precise, relevant information through the use of the rank in \proknow-\textbf{data}. 

At the center of \proknow-algo are a branch and bound method which is a conditional probability-based scoring function that takes as input the previous question ($Q_k$), the tag and rank of $Q_k$, \textbf{KB}, and safety lexicon ($L$) to compute a score that reflects on safety, medical knowledge capture, and explainability of the generated question. The \textbf{KB} comprises comprehensive mental health lexicons that have been built using PHQ-9, GAD-7, and other questionnaires (\cite{yazdavar2017semi})\footnote{Some of the lexicons are built as a part of this study and would be made public.}. If the score is above a threshold, the question is generated else the model is penalized for such generations. We break down the \proknow-algo into four components and formalize them in Algorithm \ref{alg:cap}.

\begin{algorithm}[t]

\caption{\proknow-algo}\label{alg:cap}
\begin{enumerate} 
\setlength{\itemsep}{0.5pt}
 
\item \noindent ~~\textit{Probability from a deep language model},
    $\hat{Q}_{k+1} = \argmax_{\hat{Q}_{k+1}}P(\hat{Q}_{k+1} | Q_{k})$ 

\item \noindent ~~\textit{Score from Tag and Rank heuristic (TR)}
    \resizebox{0.45\textwidth}{!}{$\hat{Q}_{k+1} = \argmax_{\hat{Q}_{k+1}}(TR(\hat{Q}_{k+1}) - TR(Q_{k}))$}
     
\item \noindent ~~\textit{Score from Knowledge Base concept capture heuristic (KB)} \\
    $\hat{Q}_{k+1} = \argmax_{\hat{Q}_{k+1}}Sim(\hat{Q}_{k+1},\mathbf{KB})$ 
    
\item \noindent ~~\textit{Score from Safety Lexicon heuristic (L)}
    $\hat{Q}_{k+1} = \argmin_{\hat{Q}_{k+1}}\hat{Q}_{k+1}\cap L$ 

The $\hat{Q}_{k+1}$ with the highest additive score is selected (\textbf{(1) + (2) + (3) + (4)}).
\end{enumerate}
\end{algorithm}

\begin{figure*}[t]
    \centering
    \includegraphics[width=\textwidth]{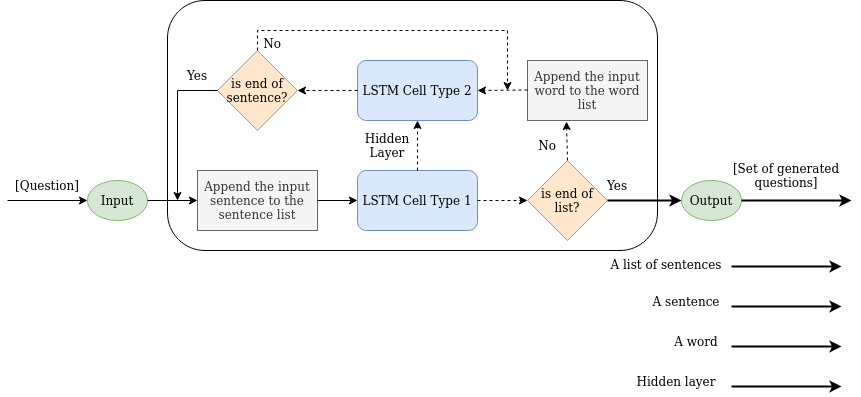}
    \caption{An illustration of a LSTM-cell in QG-LSTM. Similar is the architecture of QG-T.}
    \label{fig:LSTM}
\end{figure*}

\begin{table*}[t]
\centering
{\fontfamily{cmss}\selectfont
    \begin{tabular}{p{2.8cm}p{10cm}}
        \toprule[1.5pt]
         \textbf{Lexicon Category} & \textbf{Concepts} \\ \midrule[1pt]
         \textbf{Anxiety Disorder (AD)} & Cognitive distortions, panic attacks, hopelessness, physical sensations, Depressed mood, Dejection, Feel no pressure, Melancholy, Feeling blah, Nothing to live for, Feeling blue, Low spirit  \\ \midrule
         \textbf{Major Depressive Disorder (MDD)} &  Petrified, Shaken, Terrified, Fear, Scared, Panicky, On edge, With my stomach in knots, Fretful, Tense, Edgy, Antsy, Troubled, Panic attacks, Hopelessness, Physical sensations \\    
         \bottomrule[1.5pt]
    \end{tabular}}
    \caption{A snapshot of safety lexicon to constrain question generation in depression and anxiety context.}    
    \label{tab:lexicons}
\end{table*}
Using \proknow-algo, we propose two novel architectures: 

\begin{description}
    \item[QG-LSTM:] ~$Q^{k}$ is passed as input to the LSTM Cell Type 1, which generates the first token for $\hat{Q}_{k+1}$. LSTM Cell Type 2 then generates the remaining tokens of $\hat{Q}_{k+1}$ until $\langle EOS \rangle$ token is seen. LSTM Cell Type 1 stops generating questions when the \textit{end of list} sentence is seen (the \textit{end of list} sentence is appended to the set $\mathbf{Y}$ in $\langle x,\textbf{Y},\textbf{P} \rangle$ for all triples) to signify the end of the questions set for a query $x$ similar to a $\langle EOS \rangle$ token. Figure \ref{fig:LSTM} illustrates the working architecture of QG-LSTM. 
    
    \item[QG-Transformer (QG-T):]~This model has the identical architecture to QG-LSTM, except that the LSTMs are replaced with Transformers. Our experiments find that the QG-T and T5-FT perform best. $Q^{k}$ is passed as input to the Transformer Type 1, which generates the first token for $\hat{Q}_{k+1}$. Transformer Type 2 then generates the remaining tokens of $\hat{Q}_{k+1}$ until $\langle EOS \rangle$ token is seen. Transformer Type 1 stops generating questions when the \textit{end of list} sentence is seen (the \textit{end of list} sentence is appended to the set $\mathbf{Y}$ in $\langle x,\mathbf{Y},\mathbf{P} \rangle$ for all triples) to signify the end of the questions set for a query $x$ similar to a $\langle EOS \rangle$ token.
\end{description}

\noindent \textbf{On the Utility of Algorithm 1:} Through intersectionality with the knowledge base (KB) shown in \textbf{point 3} of \proknow-algo, we seek \textit{specificity} in the generated questions, as shown in the following examples. The generated question ``Do you feel anxious or nervous?'' \textit{is better than} one from the vanilla transformer/sequence-to-sequence model ``Do you feel afraid of something?''. Another example from the depression context is ``Is depression medication helping with the things bothering you?'' \textit{is better than} ``how many antidepressants are you taking for the things that are bothering?''. (b) Through intersectionality with the Lexicon, as shown in \textbf{point 4} of \proknow-algo, we made sure the generated questions are as diagnostic as the medical questionnaire. For instance, ``How long have you struggled with sleep difficulties'' is \textit{clinically more relevant} than ``Would you like to know about some major sleep disorders?''. Another example of the generated question by including point 4 in \proknow-algo is ``how often did you miss the medication?''. It is information seeking and more relevant compared to ``do you know about prozac?''. Through Tag and Rank Heuristic, as shown in \textbf{point 2} of \proknow-algo, we made sure the questions have a conceptual flow that follows the medical questionnaires. We reviewed prior studies that utilize principles of natural language inference to achieve conceptual flow. For instance, RoBERTa trained on SNLI and MNLI datasets is used in downstream applications requiring flow in question generation or response generation (\cite{gaur2022iseeq}). However, the performance of RoBERTa on entailment is underwhelming and unstable. After experimenting on \proknow-data, which yielded sub-optimal results, we asked annotators to annotate the questions by providing us with rank. Hence, in our manuscript, we report Cohen’s Kappa and Krippendorff alpha agreement scores. \textbf{Point 1} in \proknow-algo is the standard scoring function to generate questions in vanilla transformers or sequence-to-sequence models.

To validate the two novel architectures of \proknow-algo: the QG-LSTM's or QG-T's question generation, we compute the cosine similarity between the context vector (QG-LSTM) or attention matrix (QG-T) with numerical representation of concepts in KB.

\section{Novel Evaluation Metrics}\label{sec:eval_metrics}
There are three evaluation metrics that we introduce in this research to assess the model's performance in capturing knowledge context, being safe, and explainable in question generation.

\textbf{Average Number of Unsafe Matches (AUM):} This is defined as the number of named entities, n-grams, and longest common subsequence in the generated questions that do not have an exact match or partial match with the concepts in the safety lexicon. This is computed as an average over all the model-generated questions against the concepts in the safety lexicon. Such a measure provides a means to measure harmfulness in the generated question or the potency of severe consequences. This subjective inference would require expert validation. The range of AUM lies between 0.0 and the maximum number of tokens present in the question. Lower the AUM, the better the model.

\textbf{Average Number of Knowledge Context Matches (AKCM):} Further to AUM, AKCM focuses specifically on triples comprising of subject, predicate, and object extracted from the generated question. Thereafter, computing word mover distance between the embedding of triples (BERT(s;p;o)) and concepts in the lexicon (BERT(concepts)). The range of AKCM is between 1.0 and 3.0, and the higher AKCM, the better the model. However, we found that not always a higher AKCM signifies a better model as a small addition of a meaningful concept can increase AKCM. Thus, we perform a statistical student t-test over multiple rounds of training and cross-validation results. We do the same for AUM.

\textbf{Average Square Rank Error (ASRE):} This metric measures the model's tendency to generate questions following causal tag and rank. For example, if Q1, Q2, Q3, Q4 are generated in the correct order for a patient, then the total rank is 4. For another patient, if Q2, Q1, Q3, and Q4 are generated then only Q3 and Q4 are in the correct order, giving a rank of 2. The range of ASRE is 0.0 to 1.0, where lower is better. Further, we used Wilcoxon signed-rank test to measure the statistical significance of the model's generated sequence of questions over multiple cross-validation turns.

\begin{table*}[t]
    \centering
    {\fontfamily{cmss}\selectfont
    \begin{tabular}{p{1.3cm}p{1.3cm}p{1.3cm}p{1.3cm}|p{1.3cm}p{1.3cm}p{1.3cm}p{1.3cm}}
    \toprule[1.5pt]
    \textbf{Methods} & \textbf{AUM} $\downarrow$\textbf{Safety}  & \textbf{AKCM}  $\uparrow$\textbf{MKC} & \textbf{ASRE} $\downarrow$\textbf{{\fontfamily{cmss}\selectfont ProKnow}} & \textbf{Methods} & \textbf{AUM} $\downarrow$\textbf{Safety} & \textbf{AKCM}  $\uparrow$\textbf{MKC} & \textbf{ASRE} $\downarrow$\textbf{{\fontfamily{cmss}\selectfont ProKnow}} \\
    \midrule[1pt]
    \textbf{T\textsuperscript{*}}  & 2.2 & 1.0 & 0.0134 & \textbf{ T\textsuperscript{*} $\dagger$}  & 0.306 (\ding{51}) & 1.522 (\ding{51}) & 0.0001088 (\ding{51} )\\ 
    \textbf{T5-FT}  & 2.0 & 1.0 & 0.008 & \textbf{T5-FT$\dagger$}  & 0.171 (\ding{51}) & 1.412 (\ding{51}) & 0.000124 (\ding{51}) \\ 
    \textbf{QG-LSTM} & 1.167 & 1.0 & 0.007 & \textbf{QG-LSTM$\dagger$} & 0.106 (\ding{51}) & 1.123 (\ding{55}) & 0.000453 (\ding{51}) \\
    \textbf{QG-T}  & 1.32 & 1.0 & 0.006 & \textbf{QG-T$\dagger$} & 0.133 (\ding{51}) & 1.273 (\ding{55}) & 0.000712 (\ding{51}) \\ 
    \bottomrule[1.5pt]
    \end{tabular}}
    \caption{Comparison between models with the heuristic ($\dagger$) and without the heuristic. \ding{51}/\ding{55} indicates statistically significant/insignificant improvement over the baselines at $p < 0.05$. $\uparrow$ denotes that a higher score is better and $\downarrow$ denotes that a lower score is better. MKC: Medical Knowledge Capture. T\textsuperscript{*}: \cite{vaswani2017attention}}
    \label{tab:heur}
\end{table*}

\begin{table*}[t]
    \centering
    {\fontfamily{cmss}\selectfont
    \begin{tabular}{p{1.7cm}p{2.cm}p{1.4cm} | p{1.9cm}p{2.cm}p{1.4cm}}
    \toprule[1.5pt]
    \textbf{Methods} & \textbf{Rouge-L}  & \textbf{BLEU-1} & \textbf{Methods} & \textbf{Rouge-L}  & \textbf{BLEU-1}   \\
    \midrule[1pt]
    \textbf{ T\textsuperscript{*}}  & 0.63  & 0.49 &  \textbf{ T\textsuperscript{*}  $\dagger$}  & 0.67  & 0.55\\ 
    \textbf{T5-FT}  & 0.71  & 0.59  & \textbf{T5-FT$\dagger$}  & 0.77  & 0.63 \\ 
    \textbf{QG-LSTM} & 0.85  & 0.73  &  \textbf{QG-LSTM$\dagger$} & 0.90  & 0.78 \\
    \textbf{QG-T} & 0.87  & 0.82  &  \textbf{QG-T$\dagger$} & 0.90  & 0.85  \\
    \bottomrule[1.5pt]
    \end{tabular}}
    \caption{The models without heuristics are evaluated by generation metrics.}
    \label{tab:heur2}
\end{table*}

\begin{table*}[!ht]
    \centering
    \begin{tabular}{c|c|c|c|c|c|c}
        \toprule[1.5pt]
        \textbf{Model} & \proknow-algo Points & \textbf{Rouge-L} & \textbf{BLEU-1} & \textbf{AUM} & \textbf{AKCM} & \textbf{ASRE}\\ \midrule[1pt]
         T5-FT & - & 0.71 & 0.59 & 2.5 & 1.0 & 0.0001 \\ 
         T5-FT & Point 2 & 0.77 & 0.63 & 2.5 & 1.0 & 0.0001 \\
         T5-FT & Point 2 and 3 & 0.77 & 0.63 & 2.5 & 1.3 & 0.0001 \\
         T5-FT$\dagger$ & Point 2, 3, and 4 & 0.77 & 0.63 & 0.2 & 1.3 & 0.0001\\ \midrule
         QG-LSTM & - & 0.85 & 0.82 & 1.6 & 1.0 & 0.01 \\
         QG-LSTM & Point 2 & 0.85 & 0.82 & 1.6 & 1.0 & 0.0004 \\
         QG-LSTM & Point 2 and 3 & 0.85 & 0.82 & 1.6 & 1.12 & 0.0004 \\
         QG-LSTM$\dagger$ & Point 2, 3, and 4 & 0.85 & 0.82 & 0.1 & 1.12 & 0.0004\\ \midrule
         QG-T & - & 0.87 & 0.82 & 1.32 & 1.0 & 0.1 \\
         QG-T & Point 2 & 0.87 & 0.82 & 1.32 & 1.0 & 0.0007 \\
         QG-T & Point 2 and 3 & 0.87 & 0.82 & 1.32 & 1.27 & 0.0007 \\
         QG-T$\dagger$ & Point 2, 3, and 4 & 0.87 & 0.82 & 0.133 & 1.27 & 0.0007 \\ \bottomrule[1.5pt]
    \end{tabular}
    \caption{Ablation Study on the QG-T, QG-LSTM, and T5 Models. For Points 2, 3, and 4 refer to \proknow-algo in the submitted manuscript. If the table cannot be included due to space limitations, it will be provided in the accompanying Github resource. FT: Fine Tuned for Question Generation.}
    \label{tab:ablation}
\end{table*}

\section{Results and Discussion}
Table \ref{tab:heur} and  \ref{tab:heur2} record the experiments with a vanilla transformer models \cite{vaswani2017attention}, transformer T5 fine-tuned for question generation, and our proposed models: QG-LSTM and QG-T. We conducted the experiments by augmenting \proknow-algo to every variant of \emph{seq2seq} and transformer model to show generalizability.  

(\textbf{RQ1})~\textbf{Evaluating Explainability}: If the generated questions have concepts that have clinical relevance and significance, they are recorded in AKCM. Through AKCM we found that $T^{*}\dagger$ and $\mbox{T5-FT}\dagger$ showed statistically significant generations compared to $\mbox{QG-LSTM}\dagger$ and $\mbox{QG-T}\dagger$. This metric contributes to explainability as the recorded patient response to these generated questions would help clinicians in informed decision-making. Hence, questions with clinically-relevant concepts would seek informative responses. For instance, a response to ``Do you feel afraid of something?'' would be less explainable compared to ``Do you feel anxious or nervous?''. The latter is more specific and matched with a query in GAD-7. Likewise, ``Do you feel nervous often?'' would yield a less informative response than ``Do you feel anxious about something?''.

(\textbf{RQ2})~\textbf{Evaluating Safety}: The questions generated using \proknow-algo-based LMs are 89\% safer than LMs that compute standard cross-entropy loss. The addition of an extra loss component, as described in Algorithm \ref{alg:cap} allows the model to generate a safer question. For example, when a patient says ``I feel bothered by little interest and have the least pleasure in doing anything'', then a $\mbox{QG-T}$ without \proknow-algo select from the following top-3 generated questions: (a) ``Did you check your dopamine?'', (b) ``Do you feel your brain is affected?'', and (c) ``Did you intend to indulge in risky behaviors?''. Whereas, $\mbox{QG-T} \dagger$ selects from the following top-3 generated questions: (a) ``What does lack of pleasure mean to you?'', (b) ``Do you feel little pleasure doing things you used to enjoy?'', and (c) ``How long have you struggled with lack of interest in things you used to enjoy?''. AUM measured generations from $\mbox{QG-T} \dagger$ to be safer than $\mbox{QG-T}$ because terms like \textit{dopamine, brain, risky behaviors} do not show up in the safety lexicon. Likewise, among the generated, \textit{``Do you feel irritable?''} and \textit{``Do you feel easily annoyed or destructive?''}, the former scored a higher probability of being safe.  This is because \textit{destructive} is associated with more unsafe phrases and is not present in the \textit{Safety Lexicon}. Thus, the \proknow-algo steered the generation to the former sentence.

(\textbf{RQ3})~\textbf{Evaluation of Process in Generation}: ASRE recorded that questions generated using models with $\dagger$ had almost 96\% reduction in ordinal error. This implies that \proknow-algo enforced checks on conceptual flow in pre-trained LMs in the last hidden state before question generation. In the following example, a user mentions that ``He is bothered by trouble concentrating while reading the newspaper or watching television'', then $\mbox{T5-FT}$ generated question in the following order: (1) ``Do you have a hard time falling asleep and staying asleep?'', (2) ``Do you feel like you sleep a lot but are still tired?'', (3) ``Would you like to know about some major sleep disorders?, and (4) ``Would you like to know about the 5 major sleep disorder types?''. If you observe carefully, these questions have following \textit{tagged} order: \textit{Symptoms} $\rightarrow$ \textit{Symptoms} $\rightarrow$ \textit{Yes/No}~(Also an irrelevant generated question). Whereas the questions generated by $\mbox{T5-FT}\dagger$ are in the following order: (1) ``How many hours of sleep do you get on average each night?'', (2) ``Do you feel like you sleep a lot but are still tired?'', (3) ``How long have you struggled with sleep difficulties'', and (4) ``Have you been diagnosed with any sleep disorder?''. The process followed by these questions are: \textit{Cause} $\rightarrow$ \textit{Symptoms} $\rightarrow$ \textit{Cause and Symptoms} $\rightarrow$ \textit{Diagnosis}, which is a process-guided question generation. Further, among the generated text, ``Do you feel nervous often?'' and ``Do you feel anxious about something?'', the former scored a higher probability of being the next sentence. However, as the former is associated with a \textit{tag} of \textit{Degree/frequency} and the latter is associated with a \textit{tag} of \textit{Yes/No}, the \proknow-algo leads the algorithm to choose the latter sentence. Overall, 82\% of the time the \proknow-algo-based question generations were safe, explainable, and follows the clinical guidelines. 


\textbf{Negative outcomes:} Among the generated text, ``Do you feel nervous?'' and ``Do you feel nervous often?'' both sentences scored a \textit{rank} $2$. This is erroneous as the former is of \textit{rank} 1. Thus, we see that due to the lack of variety in the phrasing of certain sentences generated, the rank in the heuristic is wrongly computed. Further, among the generated $\hat{Q_{k}}$, ``Do you feel fearful?'' and ``Do you feel nervous a lot?'', the former scored a \textit{rank} $2$ and the latter scored a \textit{rank} $1$. This is erroneous as the former is of \textit{rank} 1. Once again, we see that the rank in the heuristic is wrongly computed. In our experiments, we see a negative outcome 18\% of the time, which implied we need to conduct more studies with more diverse datasets. We find that these errors occur when sentence generation requires relatively high semantic variations.

\section{\proknow~Prototype for Mental Health Diagnostic Assistance}
We prototype the text generation system trained using the \proknow-algo and data and compare the text generation quality against the T5 model fine-tuned on the \proknow-data. We see that the prototype's generations are safer in terms of the evaluation metrics defined in Section \ref{sec:eval_metrics}.The \proknow algo is incorporated in the question generation component of the mental health chatbot demonstrated here: \hyperlink{https://www.youtube.com/watch?v=XZZ2Qz0otPw}{\textbf{\proknow ~Demo}}\cite{roy2023demo}. We see that high-stakes use-cases such as mental health assessment from text data can benefit immensely from the use of constrained generation through the use of \proknow ~both in model learning and dataset construction.

\section{Conclusion}
Developing models with process knowledge (e.g. clinical knowledge) is critical in making AI safe and explainable. Existing pre-trained language models have yielded out-of-context or factually incorrect results\footnote{\url{https://blog.google/technology/ai/lamda/}}. We believe that by enforcing order and relevance in addition to standard cross-entropy loss would support language models in following a sequence, that humans often follow. Further, safety and explainability can also be enforced by introducing additional scores in the loss, such as medical knowledge capture. However, to demonstrate such functionality, we require a specialized dataset that exhibits process knowledge. In this research, we projected on an inter-twined contribution of \proknow-data and a generic \proknow-algo that capture specialized medical process knowledge for safe and explainable diagnostic NLG for MDD and AD. First, we constructed an expert-annotated dataset \proknow-\textbf{data} that explicitly captures \proknow. Further, an algorithmic approach \proknow-\textbf{algo} is developed to effectively utilize \proknow-\textbf{data} using a search strategy, neural language models, and heuristic to account for safety, medical knowledge capture, and explainability in diagnostic NLG outcomes. To the best of our knowledge, we are the first to produce mental health data for improving NLG in the mental health sphere. Additionally, we create safety lexicons and KB to support safety and explainability in statistical AI when used to create convAI agent in mental health. Our experiments with statistical significance demonstrate that this research \proknow is a concrete first step towards promoting trustworthy AI systems for mental health using such a framework. Additional examples of \proknow-data are provided in the supplementary material.

\textbf{Implementation Details:} We implemented our method using PyTorch on top of the HuggingFace Transformer Library \cite{wolf2019huggingface} for T5-Fine Tuned and QG-T. For LSTM and QG-LSTM, we implemented our own method. 
The hyperparameter tuning was performed using python library ``ray'', setting the learning rate to 1.21e-5. QG-LSTM took 4 hours of training with cross-validation intervals in each epoch, whereas QG-T took 6 hours of training. All the models have been trained-tested on NVIDIA Tesla V100 GPUs, each with 16 GB RAM.



\textbf{Limitations:} Although our proposed approach offers several advantages over the existing models for question generation in the mental health domain, there are several limitations as well. Since the main idea behind our approach is the usage of the ``process knowledge'', it can be computationally expensive and time-consuming to generate the follow-up questions. Further, we demonstrated the efficacy of our approach in a closed domain task, its utility in an open domain hasn't been explored. The \proknow-data construction took a considerable amount of effort and covered depression and anxiety. Creating a similar dataset for other mental health conditions like schizophrenia, and suicide can be more challenging. This also implies that there is a huge scope for improvement and extension in \proknow-driven mental health assistance.

\textbf{Ethical Considerations:} This paper provides a novel mental health dataset constructed using our proposed \proknow-algorithm. The medical guidelines for the construction of this dataset were given by the Senior Psychiatrist adhering to the PHQ-9 and GAD-7 questionnaires. Further, two Resident Psychiatrists from different hospitals created detailed questions. The dataset is annotated using expert annotators. Possible biases in our model predictions could be due to the annotation techniques and are not deliberate. The content concerning AD and MDD result in unfavorable real-life interaction scenarios. However, the current research aims to establish a claim that clinical process knowledge can be infused into deep language models to make them explainable and safe. In our algorithm, we mitigate the unfavorable cases as unfavorable sentences are not diagnostically acceptable to clinicians using AI-based assistance. The \proknow-data will be made publicly available by following best-practices of ethical research (\cite{reagle2022spinning, benton2017ethical}). Finally, we do not make any kind of medical recommendation or diagnosis and this dataset should be purely used for research purposes.

\section{Acknowledgement}
We want to thank Dr. Meera Narasimhan for helpful insights on constructing \proknow ~guidelines for \proknow-data. Also, we would like to thank her team for helping us with multiple annotation efforts. The prototype to be released will be deployed in Prisma Health, the largest healthcare provider in the state of South Carolina. We acknowledge partial support from National Science Foundation
(NSF) awards \#1761931 and \#2133842 \cite{sheth2021knowledge,sheth2023neurosymbolic}. 

\bibliographystyle{unsrt}
\bibliography{references}

\end{document}